\documentclass[letterpaper, 10 pt, conference]{ieeeconf}

\IEEEoverridecommandlockouts
\makeatletter\let\proof\@undefined\let\endproof\@undefined\makeatother

\usepackage{cite}
\usepackage{array,multirow}
\usepackage{tabularx,booktabs}
\usepackage{algorithm}
\usepackage[noend]{algpseudocode}
\usepackage{lipsum}
\usepackage{amsmath}
\DeclareMathOperator*{\argmax}{argmax}
\usepackage{amsthm, amssymb}
\usepackage{amsfonts}
\usepackage{tabularx,booktabs}
\usepackage{graphicx}
\usepackage{color, soul}
\usepackage{subfig}
\usepackage{soul}
\usepackage[font={small}]{caption}

\newtheorem{assump}{Assumption}

\usepackage{url}

\begin{document}

\title{\LARGE \bf
An Efficient Scheduling Algorithm for Multi-Robot Task Allocation in Assembling Aircraft Structures
}

\author{Veniamin~Tereshchuk$^{1}$, 
		John~Stewart$^{1}$, 
        Nikolay~Bykov$^{1}$, 
        Samuel~Pedigo$^{2}$,   
        Santosh~Devasia$^{1}$ 
        and Ashis~G.~Banerjee$^{3}$
\thanks{This work was made possible by the support of Boeing Research and Technology BR\&T-1218-282 and the Boeing Advanced Research Center (BARC) at the University of Washington, Seattle, WA 98195, USA.}%
\thanks{$^{1}$V. Tereshchuk, J. Stewart, N. Bykov, and S. Devasia are with the Department of Mechanical Engineering, University of Washington, Seattle, WA 98195, USA {\tt\small bter, austin95, nikbyk,  devasia@uw.edu}}%
\thanks{$^{2}$S. Pedigo is with The Boeing Company, Everett, WA 98203, USA {\tt\small samuel.f.pedigo@boeing.com}}%
\thanks{$^{3}$A. G. Banerjee is with the Department of Mechanical Engineering and the Department of Industrial \& Systems Engineering, University of Washington, Seattle, WA 98195, USA {\tt\small ashisb@uw.edu}}%
}

\maketitle

\begin{abstract}
Efficient utilization of cooperating robots in the assembly of aircraft structures relies on balancing the workload of the robots and ensuring collision-free scheduling. We cast this problem as that of allocating a large number of repetitive assembly tasks, such as drilling holes and installing fasteners, among multiple robots. Such task allocation is often formulated as a Traveling Salesman Problem (TSP), which is NP-hard, implying that computing an exactly optimal solution is computationally prohibitive for real-world applications. The problem complexity is further exacerbated by intermittent robot failures necessitating real-time task reallocation. In this letter, we present an efficient method that exploits workpart geometry and problem structure to initially generate balanced and conflict-free robot schedules under nominal conditions. Subsequently, we deal with the failures by allowing the robots to first complete their nominal schedules and then employing a market-based optimizer to allocate the leftover tasks. Results show an improvement of 11.5\% in schedule efficiency as compared to an optimized greedy multi-agent scheduler on a four robot system, which is especially promising for aircraft assembly processes that take many hours to complete. Moreover, the computation times are similar and small, typically hundreds of milliseconds.
\end{abstract}

\begin{keywords}
Industrial Robots, Intelligent and Flexible Manufacturing, Planning, Scheduling and Coordination
\end{keywords}

\section{Introduction}

Aircraft structure assembly is a large-scale manufacturing process with many repeated subprocesses such as drilling, fastening, trimming, and painting that naturally lend themselves to automation. Using multiple robots to work cooperatively offers a means of increasing the production speed, and hence, multi-robot systems (MRSs) have been widely studied in manufacturing applications \cite{mult_obj_heur_2017, GA_fms_2012, sim_agv_2018}. Here, we consider the  efficient scheduling of such an MRS (comprising multiple 
stationary robotic manipulators with limited but overlapping reach) during the assembly of large aircraft structures such as the wing and fuselage.

To maximize the production output, the MRS must make efficient utilization of all the robots by minimizing their idle times and travel costs \cite{K_means_2011}. For stationary robots, the idle time is governed by uneven workload distribution among the robots. Therefore, suitable task allocation is of primary concern, followed by appropriate task scheduling for minimal travel cost. The major challenges are balanced task assignment and efficient, collision-free scheduling of the assigned tasks. Another  challenge is to manage intermittent failures caused by robot calibration and communication errors, or end effector breakdowns that require a robot to be pulled out of operation for repairs. This typically necessitates real-time reallocation of these tasks \cite{rt_auction_2005}. Hence, the MRS  efficiency depends also on fast computation and the ability to avoid  frequent rescheduling.

In this letter, we present a scheduling method that leverages the structure of the problem and workpart geometry to allocate the tasks among multiple robots. Specifically, we develop a dual-stage scheduling approach, where a conflict-free, balanced nominal schedule is first generated that avoids the need for immediate rescheduling after failure instances. In the second stage, the tasks leftover during the failures are reallocated using a market-based optimizer to enable efficient, collision-free cooperation. We test our method on a wing skin attachment problem involving drilling of approximately 2000 holes by four robotic arms, and report very promising results. Therefore, our main contribution is a fast MRS scheduling method that is useful for a variety of repetitive aerospace manufacturing processes.

\section{Related Work}

Multi-robot task allocation (MRTA) is often formulated as a multiple traveling salesman problem (mTSP) \cite{MRTA_SOTA_2015}, a variant of the NP-hard traveling salesman problem (TSP), for which the branch and bound paradigm is commonly used to generate optimal solutions \cite{branch_bound_1997}. Large scale TSPs are computationally intractable, and, consequently, popular solution methods employ heuristics-based approximations of linear programs \cite{MRTA_Gini_2017} and mixed integer linear programs (MILPs) \cite{RAL_2016_mix_int}. For example, a Monte Carlo tree search-based heuristic was used in conjunction with the branch and bound algorithm to yield near optimal robot task allocation for 100 tasks within an hour \cite{monte_carlo_2016}. A MILP approach was used in \cite{tercio_2013} to experimentally produce approximately optimal schedules for 10 robots and 500 tasks in 20 seconds, and also in \cite{RAL_2016_mix_int} for near-optimal mobile robot task planning in less than 100 seconds. Although these approximate optimization methods provide reduced computation times, fast near-optimal task allocation remains challenging for a large number of tasks. Hence, they are not particularly suitable for our problem with a large number of tasks and multiple robots.

Market or auction based methods, which are somewhat less computationally demanding, are commonly used for decentralized/distributed MRTA, especially where inter-agent communication and consensus are limited \cite{RAL_2018_consensus}, \cite{cbaa_2008}. These methods have seen significant success in multiple mobile robot system (MMRS) task allocation, especially for unknown region exploration \cite{mb_survey_2006,rt_auction_2005,UAV_2005,UAV_2007}. Auction based approaches are well suited for dynamic tasks and environments, and benefit from the ability to distribute the computational burden among the agents. However, they often produce less optimal solutions than the fully centralized optimization approaches \cite{MRTA_SOTA_2015}. Recent efforts have focused on improving the optimality of the solutions. For example, Korsah et. al. \cite{seed_korsah_2011} seeded their 
market-based method with pre-computed optimal schedules of static tasks, which, however, could be computationally expensive for large-scale problems. Task cluster allocation was also shown to have promising results in improving the workload balance in \cite{K_means_2011}. However, it also led to increased computational complexity due to the necessity to bid on all the cluster combinations. Therefore, similar to the centralized optimization methods, the main challenge is that improved solution quality of market-based methods often comes at the cost of increased computation times for large scale problems as in the current application. 

We address the trade-off between quality and computation time for large-scale problems by employing a market-based method as a schedule refinement technique on a smaller scale problem defined by the leftover tasks, instead of using it to directly solve the full scheduling problem. In particular, our method exploits a schedule placement technique similar to the one used in \cite{TeSSi_2015}, where optimal placement of the task within the schedule helped inform bids and increased solution optimality. 

We also use a graph-partitioning method, commonly used in high performance computing research, for nominal scheduling. Partitioning techniques are not as commonly used for MRTA as centralized optimization or market-based approaches, but they are shown to have direct benefits in solving task allocation problems with favorable scaling properties in \cite{partition_2012}. Moreover, partitioning has shown effectiveness in fair global task subdivision for heterogeneous robot teams \cite{fair_subdiv_2013}, as well as enabling quicker convergence for workload division in multi-robot exploration \cite{partition_2015}. Our work extends these partitioning methods to the structured nature of aircraft assembly problems for equitable work distribution. We build upon these methods by using them not only for equal task allocation, but also to enable de-conflicted task scheduling.

\section{Problem Formulation}

We consider the class of manufacturing problems where $m$ pairs of homogeneous, stationary robots $r_i$ perform $n$ discrete assembly tasks on a workpart, where $n>>m$. Since task allocation of any robot inherently depends on the schedules of all the other proximate robots in order to avoid collisions, this class of problems best fits in the single-task robots, single-robot tasks, time extended assignment with cross-schedule dependencies (XD[ST-SR-TA]) category of problems in the MRTA taxonomy \cite{Korsah_Tax}. The robots are situated about the part along opposing pairs. The reach of each robot is limited to a subset of tasks but overlaps with the reach of the adjacent and opposite robots. This arrangement of the robots, shown in Fig. \ref{generic_prob_full}, defines an axial direction that is leveraged for task scheduling in the next section.

Aircraft assembly often has variability in its tasks from airplane to airplane that arises from components that are missing or intentionally omitted from the part. The definitions of which tasks need to be completed on a given part are referred as the condition of assembly (COA). Schedules need to be generated for these different COAs. Additionally, the robots experience intermittent failures that require in-process repairs. These failures need to be addressed by the scheduling approach to preclude collisions or excessive idling.

\subsection{Schedule Efficiency Metric}
The MRS schedule-efficiency metric $\epsilon$ is 
defined as
\begin{equation}\label{efficiency1}
	\epsilon = \frac{t_{min}}{t_{act}}
\end{equation}
where $t_{min}$ is the minimum possible time for all the $2m$ robots to complete all the $n$ tasks and $t_{act}$ is the time that the robots actually take. We assign $t_{min}$ to be the sum of the service times $t_{s,j}$ (the time required to complete the $j$-th task) of all the $n$ tasks and the sum of the failure times $t_f^i$ for all $2m$ robots, split evenly among all the robots. Repair time does not penalize efficiency as the robots are not capable of doing work during this time. Therefore, the efficiency metric in (\ref{efficiency1}) becomes 
\begin{equation}\label{T_min}
	t_{min} = \frac{\sum\limits_{j=1}^{n} t_{s,j} + \sum\limits_{i=1}^{2m} \text{$t_f^i$}}{2m} \quad .
\end{equation}

For the current application, travel time between the tasks is significantly less than the task
completion time. So, we include an approximated travel time in each $t_{s,j}$ and do not consider it separately in our formulation. We use $t_{{act}}$ to denote the time the last-to-finish robot takes to complete its allocated tasks, which is the maximum sum of the makespan $t_s^i$ (the time for robot $i$ to complete all its assigned tasks) and its repair time $t_f^i$, expressed as,
\begin{equation}\label{T_act}
    t_{act} = \max_{i} \text{($t_s^i + t_f^i$)} \quad .
\end{equation}

Combining (\ref{T_min}) and (\ref{T_act}), the efficiency metric  for a task schedule is expressed as
\begin{equation}\label{efficiency2}
	\epsilon = \frac{\sum\limits_{j=1}^{n} t_{s,j} + \sum\limits_{i=1}^{2m} \text{$t_f^i$}}{2m \max(\text{$t_s^i + t_f^i$})} \quad .
\end{equation}

\section{Conflict-Free Nominal Scheduling}

This section outlines the proposed nominal scheduling method. We first show the existence of a solution for a uniform task distribution, and then show robustness of the solution due to COA variations by relaxing the uniformity assumptions. For $m$ opposing robot pairs, we designate the makespan for the $i$-th pair as $t_p^i$, and further distinguish the top and bottom robot makespans as $t_t^i$ and $t_b^i$, respectively. 

\subsection{Fair Partitioning}

Our method makes the following three assumptions about task distribution uniformity, which are typical for aerospace assembly applications. 

\begin{assump}\label{Assumption1}
The task distribution uniformity allows the workpart to be divided into $m$ regions along the axial direction with equal service time $t_p$, each assigned to a robot pair, assuming no failure occurrence.
\end{assump}

\begin{assump}\label{Assumption2}
Each opposing robot pair is placed such that the tasks in its assigned region can be assigned evenly between the two robots into top and bottom regions (with task equal task service times $t_t^i = t_b^i = 0.5 t_p$)  as illustrated in Figure 1.
\end{assump}

\begin{assump}\label{Assumption3}
As the robots complete their assigned tasks, they traverse the workpart in the axial direction at equal rates $q$.
\end{assump}

\begin{figure}[thpb]
      \centering
      \includegraphics[width=.95\columnwidth]{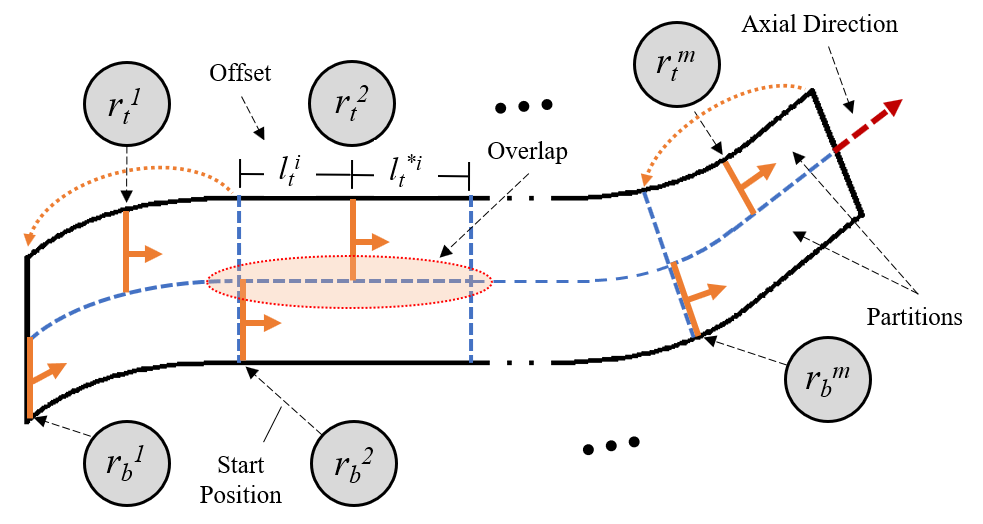}
      \caption{A generalized aircraft structure that has tasks (nearly) uniformly distributed across the work part. The part is assembled by $m$ robot pairs arranged along the axial direction of the work part.
      Partitioning allocates the tasks among all the $2m$ robots, such that the service time of each partition is equal and idling is minimized. The overlap region contains tasks that can be serviced by more than one robot, and can, therefore, be reallocated to balance the workload of the robots. Sequencing the tasks in each partition in the axial direction and offsetting the start locations of one side of robots (top robots) enables collision-free operation. Once the top robots reach the end of their partitions, they return to the start of their partitions, and service the remaining tasks in the same axially sequenced manner. 
      The bottom robots have moved out of the way at this point, and no collisions occur.
      }
      \label{generic_prob_full}
   \end{figure}

By Assumptions \ref{Assumption1} and \ref{Assumption2}, we use partitioning to achieve equitable task allocation among all the robots and minimal idle cost, resulting in $2m$ partitions, each one assigned to a different robot. We employ a pairwise optimization method as in \cite{fair_subdiv_2013} for partitioning. An example partitioning of the work part is shown in Figure \ref{generic_prob_full}.

\subsection{De-conflicted Offset Scheduling}

The  partitioned work is scheduled by sequencing all the tasks in every partition in the same axial direction and offset the start locations of all the robots on one side (top or bottom) of the work part as shown in Figure \ref{generic_prob_full}\footnote{This is feasible only if the task ordering constraints do not prevent axial sequencing.}. We denote the $i$-th offset and non-offset robots as top $r_t^i$ and bottom $r_b^i$ robots (as in Figure \ref{generic_prob_full}), respectively, but either the top or bottom robot can be offset in practice. The robots $r_t^i$ with offset start locations return to finish the rest of the work in their respective partitions using the same axial sequencing when they reach the partitions' ends, as shown in Figure \ref{generic_prob_full}. For the $i$-th offset robot, $r_t^i$, we designate the offset location as a distance $l_t^i$ from the start of its partition, which defines a distance $l_t^{*i}$ to the end of its partition, as in Figure \ref{generic_prob_full}. These have traversal times (time to complete the work) $t_t^i$ and $t_t^{*i}$, respectively.

Our first claim is that the offset-based task scheduling, subject to the following two constraints, ensures safety, i.e., collision-free operation. Safety is specified by the spatial proximity constraint that requires the distance between the end-effector positions (task locations) of any two robots to be  at least  $\alpha d_{ee}$ at all times, where  $d_{ee}$ is the diameter of the end-effectors and $\alpha \geq 1$ is a safety factor.\\

\textbf{Constraint 1:} We constrain the robot spacing to be sufficiently large such that the offset distances $l_t^i$ and $l_t^{*i}$ are larger than the proximity constraint: 
\begin{equation}\label{const_1}
	\text{$l_t^i$} > \alpha d_{ee}
	\quad\mathrm{and}\quad
	\text{$l_t^{*i}$} > \alpha d_{ee} \quad .
\end{equation}
\textbf{Constraint 2:} We constrain the offset of robot $r_t^i$ such that the time it takes to reach the end of its partition, $t_t^{*i}$, is greater than that of $r_t^{i-1}$, for $i > 1$:
\begin{equation}\label{const_2}
    \text{$t_t^{*i} > t_t^{*i-1}$} \quad .
\end{equation}

The first constraint ensures that each robot has sufficient spatial separation from its opposite and adjacent robots at the start of the operation. Note that this constraint also implicitly limits the length of any partition to be no less than $2\alpha d_{ee}$, and, consequently, limits robot density. The second constraint ensures that $r_t^{i-1}$ returns to the beginning of its partition before $r_t^i$ does, which precludes collision when $r_t^i$ returns. These two constraints together allow collision-free operation because Assumption \ref{Assumption3} ensures that the robots have equal axial traversal rates $q$, and, thus, the spatial proximity constraint is not violated. 

In practice, variable task service times and COAs with missing tasks can cause the axial traversal rate of a robot to fluctuate during operations requiring the solution to be robust. Our second claim is that Assumption \ref{Assumption3} can be relaxed to allow for variation $\delta_q$ in the axial traversal rate $q$ and still preserve the collision-free nature of the nominal schedule, provided that the maximum variation $\delta_q$ in the traversal rate $q$ is small, and the constraint inequalities in (\ref{dist1}) and (\ref{dist2}) are satisfied. This follows from the fact that the distance $d_{b^i, t^{i-1}}(t)$ between the $i$-th non-offset  robot $r_b^i$ and the previous offset robot $r_t^{i-1}$ can decrease at the maximum by $2 \delta_q t$ in time $t$ where $2 \delta_q$ is the maximum difference in the traversal rates of the two robots. Then, collision between robots $r_b^i$ and $r_t^{i-1}$ can be avoided (before robot $r_t^{i-1}$ returns to the beginning of its partition) by ensuring that the distance $d_{b^i, t^{i-1}}(t)$ is sufficiently large, i.e.,
\begin{equation}\label{dist1}
 \text{$d_{b^i, t^{i-1}}(t) 
 = l_t^{*i-1}$} - 2\delta_q t  > \alpha d_{ee}
\end{equation}
where the offset $l_t^{*i-1}$ is the initial distance between them. Similarly,  collision between the $i$-th non-offset robot $r_b^i$ and the next offset robot $r_t^i$ is avoided (before $r_t^i$ returns to the beginning of its partition) 
provided 
\begin{equation}\label{dist2}
\text{$d_{b^i, t^{i}}(t) = l_t^i$} - 2\delta_q t > \alpha d_{ee}, 
   \quad t  \le  \text{$t_t^{*i}$} \quad .
\end{equation}

We  also avoid collision between robots $r_t^i$ and $r_b^i$ after $r_t^i$ returns to the beginning of its partition, as in Figure \ref{generic_prob_full}. When robot $r_t^i$ returns, the distance $d_{b^i, t^{i}}$ between the two robots, which is initially at least $(q-\delta_q) t_t^{*i}$, needs to satisfy 
\begin{equation}\label{dist3}
   \text{$d_{b^i, t^{i}}(t)$} = 
   (q-\delta_q) \text{$t_t^{*i}$} - 2\delta_q(t-\text{$t_t^{*i}$}) > \alpha d_{ee} 
\end{equation}
for time $t  >  t_t^{*i}$. If the variation $\delta_q$ in the axial traversal rate $q$ is  small, then the inequalities in (\ref{dist1}), (\ref{dist2}), and (\ref{dist3}) are satisfied 
because (i)~their left hand sides approach 
the spacing between the robots ($l_t^{*i-1}$, $l_t^i$ and $l_t^{*i}$), which are sufficiently large from Constraint 1; (ii)~the time $t$ is finite and  bounded by the maximum makespan; and (iii)~the minimal distance $(q-\delta_q) t_t^{*i}$ approaches $l_t^{*i}$. Thus, the above analysis shows the existence of a collision-free nominal schedule if the maximal variation $\delta_q$ in the axial task execution rate $q$ (due to COA variation) is not too large.

\subsection{Schedule Execution and Failure Handling}
To maintain the collision-free nature of the nominal schedule and avoid frequent rescheduling, a robot returning from maintenance after a failure occurrence returns to the place in its nominal schedule where it would have been had no failure occurred. This requires the robot to skip some of its work to be dealt with after the nominal schedule is completed. These leftover tasks are then reallocated among the robots, and are executed after the nominal schedule is completed. 
\section{Leftover Scheduling}

In this section, we present the leftover scheduling method and task reallocation algorithm. For workload rebalancing to be possible, we guarantee that all the robots have leftover tasks in areas of overlapping reach with at least one other robot by omitting a subset of tasks in the region of overlap from the nominal scheduling and relegating them to the leftover scheduling. Our approach first constructs a conflict-free leftover schedule that is not necessarily balanced, and then uses a market-based algorithm to balance the workload and optimize the schedule. It is necessary for the initial schedule to be conflict-free because we can then constrain the algorithm to reallocate the tasks without introducing collisions. To simplify the construction and refinement of the leftover schedule, adjacent tasks are grouped into larger portions of work, referred to as cities (as in the TSP formulation), although this grouping is not required for the algorithm to work. The reduction of problem complexity decreases computation time (scheduling cities rather than individual tasks),
and is discussed more in Section VI. 

\subsection{Initial Leftover Scheduling}

\begin{figure}[thpb]
      \centering
      \includegraphics[width=.4\columnwidth]{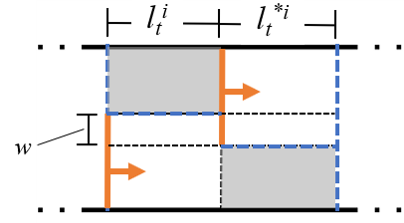}
      \caption{
      Extending the  partitions to cover an overlap region (containing tasks intentionally left out from the nominal schedule and relegated to the leftover schedule) to ensure the maximum deviation $\delta_{q,L}$ in the axial traversal rate  $q_L$ is  sufficiently small in the extended regions (shown in white) even in the presence of leftovers. The reduction of traversal rate variations along with sufficient sizing of the overlap region ensures that the initial leftover schedule is collision-free.
      }
      \label{gen_LO_IC}
   \end{figure}

We construct the initial scheduling by modifying the partitioning and offset-scheduling method used for nominal scheduling to ensure that the decreased uniformity in leftovers does not cause collisions. For this purpose, we use the overlap region between the robot pairs with width $w > \alpha d_{ee}$, and extend part of the robots’ partitions over the width of this region, the part of the $i$-th bottom robot's $r_b^i$'s partition defined by the offset $l_t^i$, and the part of robot $r_t^i$'s partition defined by $l_t^{*i}$, as shown in Figure \ref{gen_LO_IC}. Note that this overlap region should contain a sufficient number of tasks relegated from the nominal schedule such that the maximum deviation $\delta_{q,L}$ in the axial traversal rate $q_L$ is sufficiently small in the extended regions even in the presence of leftovers. 

We claim that the offset-based scheduling method results in a collision-free initial leftover schedule, provided: (i)~the robots are able to reach the tasks in the overlap region; (ii)~the spacing constraint  in (\ref{const_1}) is satisfied; (iii)~the time $t_b^i$ that robot $r_b^i$ takes to service the extended part of its partition is not more than the time $t_t^{*i}$ that robot $r_t^i$ takes to service its extended part, i.e., $t_t^{*i} = t_b^i$; and (iv) the partitioning is such that the service times of the extended parts increase along axial direction, i.e., $t_b^i < t_b^{i+1}$ to ensure that the constraint in (\ref{const_2}) is satisfied. The claim follows since both the robots, $r_t^i$ and $r_b^i$, do not move to their second non-extended regions (shown as the gray regions in Figure \ref{gen_LO_IC}) before they complete the tasks in their extended regions. Note that the distance between the robots in the extended regions can be shown to remain sufficiently large, using arguments similar to those in inequalities (\ref{dist1}) and (\ref{dist2}), provided the 
variation in the axial traversal rate is  sufficiently small. Moreover collisions are avoided when the robots are in the gray (non-extended) regions since the overlap distance is larger than $w > \alpha d_{ee}$.

%
\subsection{Market-Based Optimization Algorithm}

The schedule optimization algorithm uses a market-based method to re-balance and reschedule the leftover cities in the initial leftover schedule in order to reduce idle cost. We have $n_c$ cities, $2m$ robots, $r_i \in \mathcal{R} = \{r_1, \hdots , r_{2m}\}$, and a corresponding schedule matrix $\textbf{S} = \left[ S_1\hdots  S_{2m} \right]^T$, where $S_i$ is the schedule of robot $r_i$. Note that $r_{2i-1} = r_b^i$ and $r_{2i} = r_t^i$ from the previous Section, since we do not need the distinction between the top and bottom robots for the market-based method. The vector $S_i = \left[s_{i,1} \hdots s_{i,l-1} \ c_j \ s_{i,l+1} \hdots s_{i,n_i} \ 0 \hdots 0 \right]$ contains cities $s_{i,l} = c_j$, denoting that city $c_j$ is serviced by robot $r_i$ and that it is the $l$-th city $r_i$ services (has placement $l$ in $S_i$); each $S_i$ contains $n_s$ elements, of which the first $n_i$ are non-zero cities. We take $n_s$ as either the smallest number of the lowest service time cities it takes to exceed an equal workload, or the number of cities in the schedule with the most cities, whichever is larger. City $c_j$ has service time $t_{s,j}$; all the $n_c$ cities must be in $\textbf{S}$ for the schedule to be complete. We define the time it takes robot $r_i$ to complete its schedule $S_i$ as $t_i = \Sigma t_{s,j}$ for all $c_j$ in $S_i$. 

In our market, robot $r_i$ is looking to sell a city $c_j$ in its schedule $S_i$ to any robot $r_{k \neq i}$. The seller is a robot who \textit{most} wants to sell one of its cities, and, likewise, the buyer is a robot who \textit{most} wants to buy that city, and an agent's desire to buy or sell is proportional to the agent's deviation from an equal workload distribution, measured by a robot's \textit{utility}. For robot $r_i$, $u_i = t_i/\Sigma t_i$, and the utility vector $\textbf{U} = [u_1 \hdots u_{2m}]$ contains the utilities of all the robots. The \textit{price} at which any robot $r_i$ wants to sell a city $c_j$ in its schedule is determined by the service time $t_{i,j}$ of $c_j$ (normalized by the maximum city service time $\max(t_{i,j})$). The price matrix $\textbf{P}$ consists of $2m$ row vectors of price, one for each robot $r_i$, and each price vector $\textbf{P}_i$ contains $n_c$ elements, where each element $p_{i,j}$ is the price of city $c_j$ if $c_j$ is in $S_i$, otherwise $p_{i,j} = -\infty$. Note that the price of a city $c_j$ is independent of its placement in $r_i$'s schedule $S_i$. The city most desired to be sold, and consequently its seller, is chosen on a basis of \textit{margin}, which we define as $m_{i,j} = (\beta * u_i) + p_{i,j}$. Here, $\beta$ is a weighting factor that is used to bias the sellers to be more concerned with the utility than price, a similar market biasing approach as used in \cite{strategic_2013} for increasing optimality. From the margin matrix $\textbf{M}$, the best seller and best city to sell is found based on the maximum margin, $i^* = \argmax_i m_{ij}$ and $j^* = \argmax_j m_{ij}$, respectively. 

Once the best seller $r_{i^*}$ and the desired city to sell $c_{j^*}$ is chosen, a set of candidate buyers is chosen on the basis of which robots have the \textit{ability} to service $c_{j^*}$. Ability $a_{k,j^*}$ in our case is determined by whether $c_{j^*}$ is in $r_k$'s reach, and is 1 if it is, 0 if it is not. Each candidate buyer is then checked to see if the maximum margin value resulting from proposed transaction, $m^{*p}$, is greater than the current $m^*$, and is disqualified from the candidate set if $m^{*p} > m^*$. For each buyer $r_k$ in the candidate buyer set, the proposed transaction (resulting in proposed utility, price and margin matrices, $\textbf{U}^p$, $\textbf{P}^p$, $\textbf{M}^p$) is evaluated by the reduction in the standard deviation of the utility vector, $\Delta_k = \sigma (\textbf{U}) - \sigma (\textbf{U}^p)$, and the best buyer is chosen to be the one whose transaction will yield the highest reduction in the standard deviation of the utility vector, $k^* = \argmax_k \Delta_k$. 
We minimize the spread in utility to minimize the imbalance in work distribution. Once the best buyer $r_{k^*}$ is chosen, $c_{j^*}$ is placed in every placement $l$ of $r_{k^*}$'s schedule $S_{k^*}$, and a minimum distance $w_l$ between any two robots at any time for the entire schedule $\textbf{S}$ is determined for each placement of $c_{j^*}$ in $S_{k^*}$. The best placement $l^*$ for $c_{j^*}$ in $S_{k^*}$ is chosen as the one with the max minimum distance, $l^* = \argmax_l w_l$. The pseudo-code for the method is provided in \textbf{Algorithm 1}.

\begin{algorithm}[ht]
 \caption{Update schedule matrix $\textbf{S}$ by selling a city such that the robots' utilities are maximally equalized}
\begin{algorithmic}[1]

\State Compute $\textbf{U}$, $\textbf{P}$, $\textbf{M}$ from $\textbf{S}$
\State $m^* = \max(m_{ij})$ 
\State $i^* = \argmax_i m_{ij}$; $j^* = \argmax_j m_{ij}$
\State $\textsc{Sale} \gets$ \textsc{False} 
 \While{$\textsc{Sale} =$ \textsc{False}}
  \State $\Delta \gets \emptyset$
  \For{each $ r_k \in \mathcal{R}\backslash r_{i^*}$}
    \If{$a_{k,j^*} = 1$ }
       \State $\textbf{S}^p = \textbf{S}$
       \State $S_{i^*}^p = \left[s_{i^*,1} \hdots s_{i^*,j^*-1} \ s_{i^*,j^*+1} \ 0 \hdots 0  \right]$
       \State $S_k^p = \left[s_{k,1} \hdots s_{k,n_k} \ c_{j^*} \ 0 \hdots 0 \right]$
       \State Compute $\textbf{U}^p$, $\textbf{M}^p$ from $\textbf{S}^p$
       \State $m^{*p} = \max(m_{i,j}^p)$
        \If{$m^{*p} < m^*$}
            \State $\Delta_k = \sigma (\textbf{U}) - \sigma (\textbf{U}^p)$; $\Delta \gets \Delta \cup \Delta_k$
        \EndIf
    \EndIf
  \EndFor
    \While{$\Delta \neq \emptyset$}  
      \State $\textbf{S}^p = \textbf{S}$; $k^* = \argmax_k \Delta_k$
      \For {$l = 1, \ldots, n_{k^*+1}$}
         \State $S_{k^*}^p = \left[s_{k^*,1} \hdots s_{k^*,l-1} \ c_{j^*} \hdots s_{k^*,n_{k^*}} \hdots 0 \right]$
         \State $w_l \gets $ min pairwise robot distance in $\textbf{S}^p$
      \EndFor
    \State $w^* = \max(w_l)$; $l^* = \argmax_l w_l$
    \If{$w^* > \alpha d_{ee}$}
        \State $S_{i^*} = S_{i^*}^p$
        \State $S_{k^*} = \left[s_{k^*,1} \hdots s_{k^*,l^*-1} \ c_{j^*} \hdots s_{k^*,n_{k^*}} \hdots 0 \right]$
        \State $\textsc{Sale} \gets$ \textsc{True}
        \State \Return $\mathbf{S}^p$
    \Else
        \State $\Delta \gets \Delta \backslash \Delta_{k^*}$
    \EndIf
    \EndWhile
  \State $m^* \gets$ next highest $\max(m_{i,j})$
  \State $i^* \gets$ next highest $\argmax_i m_{ij}$ 
  \State $j^* \gets$ next highest $\argmax_j m_{ij}$ 
  \If{$m^* - p_{i^*,j^*} \leq \mu(\textbf{U}) $}
    \State $\textsc{Sale} \gets$ \textsc{True} \Comment{Sale is not useful}
    \State \Return $\textbf{S}$
  \EndIf
\EndWhile
 
 \end{algorithmic}
\end{algorithm}

If the largest minimum distance $w^* = max(w_l)$ is greater than the collision threshold, $r_{i^*}$ sells $c_{j^*}$ to $r_{k^*}$ and inserts $c_{j^*}$ in placement $l^*$, yielding the maximum geometric separation of the robots during the whole schedule. If $w^*$ is less than the collision threshold, we choose the next best buyer (based on next highest $\Delta_k$), and proceed to check all the placements of $c_{j^*}$ in the new buyer's schedule. This ensures no collisions are introduced into the schedule. If no suitable buyers are available for $c_{j^*}$, the next highest margin $m_{i,j}$ is found (resulting in a new city and seller), and the process for selling the city is repeated. This is done until a city is sold, or until all the margins, $m^* - p_{i^*,j^*} \leq \mu(\textbf{U})$, are chosen and checked for potential buyers\footnote{This guarantees the convergence of Algorithm 1.}. If no sale happens, no further optimization is possible. We run this algorithm on the initial leftover schedule.

\section{Implementation and Results}

To demonstrate the performance of our method, we present experimental results for an aircraft wing hole drilling problem using various COAs and failure cases. Since generating good-quality schedules quickly is especially important, we evaluate the method's performance on the basis of schedule efficiency and computation time. We also compare our method with an optimized greedy scheduler, as in \cite{TeSSi_2015}, that uses the same scheduling framework (nominal scheduling and leftover task rescheduling) and leftover optimization as our method. 

\subsection{Implementation}
\begin{figure}[thpb]
      \centering
      \includegraphics[width=.9\columnwidth]{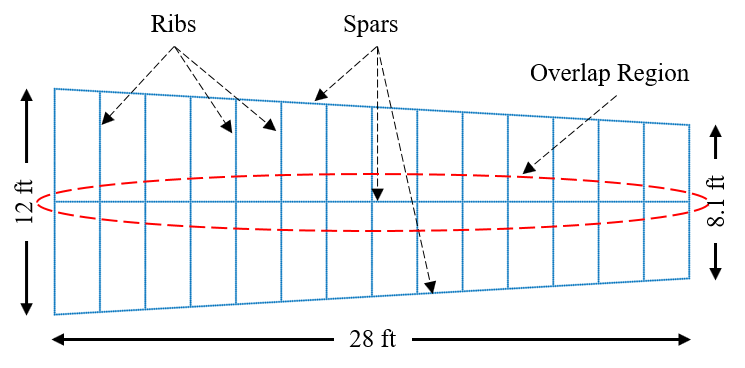}
      \caption{Mock wing used for the testing of our methods. Rows of holes to be drilled are defined by 15 ribs and three spars. Ribs have even, two-foot spacing. The middle spar, highlighted in red, offers regions where all the robots are able to service tasks in the workload re-balancing stage.}
      \label{wing_def}
   \end{figure}

We considered the problem of drilling rows of holes in a section of an aircraft wing box by four robots situated around the wing. To drill a single hole, a robot moves its end effector to the hole's location, and remains stationary for the duration of the hole's drill time as the end effector performs the drilling operation. We used an example aircraft wing with rows of evenly spaced holes defined by 15 ribs and three spars. Each spar consisted of 266 holes, while the ribs contained 109, 107, 105, 101, 99, 95, 93, 91, 87, 85, 83, 79, 77, 73, and 71 holes, respectively. Rib hole drill (service) time was set to 30 s per hole for the largest rib, and then decremented by 0.5 s for each successive rib (e.g., 29.5 s per hole for the next largest rib, 29 s per hole for the one after that, and so on). Every spar section between any two adjacent ribs was assigned the same hole drill time as that of the longer rib. We used $\alpha = 1.5$ and $\beta = 20$ as the parameters in our methods, and $d_{ee}$ was 2 ft. The offset lengths ($l_{t}^{*i}$ and $l_t^{i}$) large enough that even when robots moved from rib to rib, the distances between the robots still satisfied the traversal rate constraints in (\ref{dist1}), (\ref{dist2}) and (\ref{dist3}).
The example wing is shown in Figure \ref{wing_def}.

\begin{figure}[thpb]
      \centering
      \includegraphics[width=.8\columnwidth]{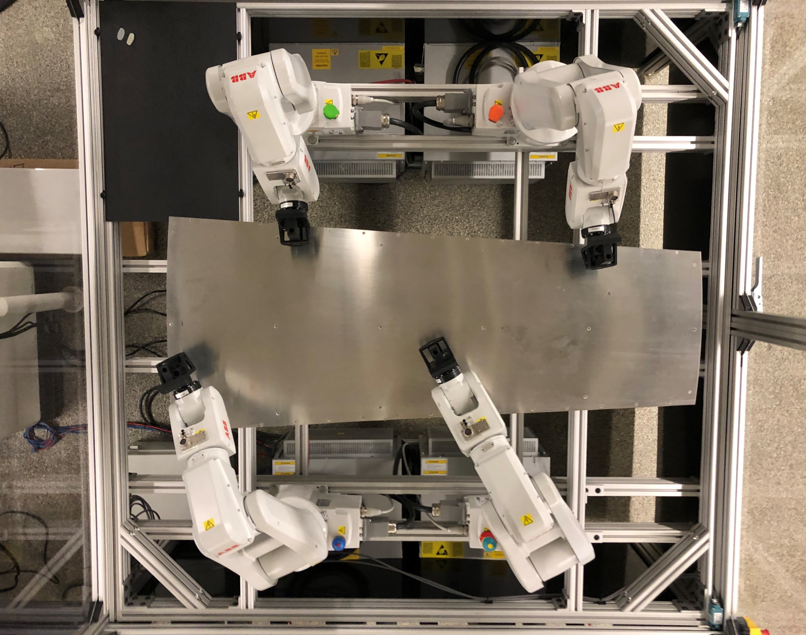}
      \caption{A 15\% reduced scale physical cell used to test the scheduling methods and validate the simulation results. The robot controllers interfaced with the cell control software, where the generated schedules where executed.}
      \label{phys_cell}
   \end{figure}

All the scheduling methods were implemented in MATLAB R2016b on a quad-core Core i7-4870HQ machine with 16 GB of RAM running Windows 10. We validated the results on a physical robot cell featuring four ABB IRB-120 arms positioned about a miniature aluminum mock wing built according to the dimensions shown in Figure \ref{wing_def}. This setup is a 15\% scale version of a feasible manufacturing setup for drilling a full-sized, real aircraft wing. The wing is raised from the floor on an aluminum T-frame that holds plexi-glass shields. The robot controllers, situated on the floor beneath the robots, are collectively directed via the cell control software. The physical cell is shown in Figure \ref{phys_cell}.

The cell control software was responsible for synchronized starting of the robots' programs, as well as keeping track of each robot's schedule and feeding hole targets to the controllers (each controller always had the next three targets). It also monitored the synchronization of the robots with each other, and determined how many holes were relegated to the leftovers for the robots returning from failures. The program was implemented in C\# using the ABB PC SDK 6.07 on a quad-core Core-i5 6400 machine with 8 GB of RAM, and communicated with the robot controllers via Ethernet UDP through a network switch. A redundant collision monitoring system used streaming robot position data to reconstruct simple 2D shapes representing the robots with the desired safety margins, and checked for intersections among the shapes. This was implemented using Python 3.7 and ran on a separate machine with the same specifications.

\subsection{Experiments}
We evaluated both the methods using five different COA cases with 100 failure instances for each COA case, leading to a total of 500 test scenarios. The COA cases were generated to have gradually more components missing from the COA (inadvertently omitted parts from the drilling process, as typically seen in the industry), with the first case being the full wing with no missing component (generic COA case). 

The failure instances were generated for each robot by randomly drawing a first occurrence time,
recurrence time, and repair time from normal distributions. The means and standard deviations of the distributions were determined  using a nominal value of how many holes a $\frac{1}{4}$ inch drill bit would last in aluminum and an average drill bit replacement time. These were taken to be 300 holes and 8 minutes, respectively.
Table \ref{fail_dist_table} outlines the distribution parameters used for the generation of failure instances. Note that first occurrence time is lower than the recurrence time, accounting for the possibility of first failures occurring earlier if drill bits are not replaced from a previous drilling operation. These parameters resulted in an average of 45 leftover cities per test scenario, each containing between two and 16 holes. We implemented the optimization algorithm to optimize by these cities, as well as by the individual holes in order to see the effects of discretization on efficiency and computation time.

\begin{table}[h]
\caption{Normal Distribution Parameters for Robot Failure Occurrence Time, Recurrence, and Repair Time}
\label{fail_dist_table}
\begin{center}
\begin{tabular}{|c||c||c|}
\hline
Parameter & $\mu$ & $\sigma$ \\
\hline
First Occurrence & 5073 s & 1602 s\\
\hline
Recurrence & 6942 s & 1068 s\\
\hline
Repair Time & 480 s & 80 s\\
\hline
\end{tabular}
\end{center}
\end{table}

\vspace{-5mm}
\subsection{Results}
\begin{figure}[thpb]
      \centering
      \includegraphics[width=.9\columnwidth]{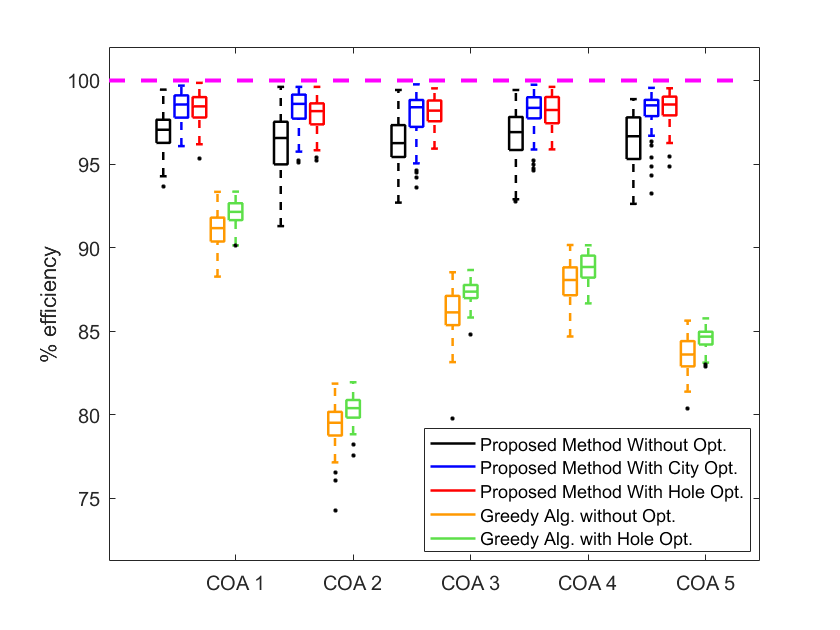}
      \caption{Comparison of the overall schedule efficiencies of our two-stage scheduling method with market-based leftover optimization (by cities and by holes), without leftover optimization, and a greedy M-TSP solver with and without leftover optimization by holes.}
      \label{efficiency_results}
   \end{figure}

{\bf{Performance improvement:~}} Figure \ref{efficiency_results} compares the efficiency of our method to that of the greedy scheduler, where we observe about 98\% efficiency on an average with our method. For every COA case, our method generates more efficient schedules than the greedy scheduler, with the efficiency increasing consistently [$F(1,998) = 3922, p = 0$] by about 
11.5\% on an average. This result is especially promising for future high-volume airplane assembly lines as 11.5\% improvement on a five-hour wing skin attachment lead time results in a 35 minute saving, or 2.75 hours over the course of one production day. We also observe that our method yields significantly more consistent schedule efficiency across the COA cases (the COA averages are within 0.5\% of each other) as compared to the greedy method (the COA averages vary by as much as 12\%). This trend indicates that our method is not affected by non-symmetric task distribution as much as the greedy scheduler.

{\bf{Impact of market-based optimization:~}} To characterize the impact of the market-based optimizer, we compare it with the partition-based scheduler without leftover optimization. We observe that the partition-based scheduler still outperforms the optimized greedy method significantly, even in the absence of optimization. This is largely due to the efficiency of the total schedules already being high because the partition-based nominal robot schedules were balanced to within the time it takes to drill a single hole. The market-based optimizer, however, consistently improves [$F(1,998) = 440.2, p = 0$] the schedule efficiency by about 2\% on an average, from 96\% to 98\%. This improvement is quite substantial when implemented across multiple processes for a high-flow aircraft production line. The results in Figure \ref{efficiency_results} are grouped by COA case\footnote{The inferior performance of the greedy method on COA 2 is a result of a rib missing from this COA, causing the robots to move toward each other, leading to long wait times in their schedules.}, and are summarized in Table \ref{results_table}.

\begin{table}[h]
\caption{Summary Statistics of Schedule Efficiencies}
\label{results_table}
\begin{center}
\begin{tabular}{|c||c||c|}
\hline
Method & Avg. Efficiency & Min. Efficiency\\
\hline
Proposed (No Opt) &  96.5\% $\pm$ 1.50\% & 91.3\%\\
\hline
Proposed (City Opt) &  98.2\% $\pm$ 1.12\% & 93.3\%\\
\hline
Proposed (Hole Opt) & 98.2\% $\pm$ 0.92\% & 94.9\%\\
\hline
Greedy (No Opt) &  85.6\% $\pm$ 4.14\% & 74.3\%\\
\hline
Greedy (Hole Opt) &  86.6\% $\pm$ 4.03\% & 77.6\%\\
\hline
\end{tabular}
\end{center}
\end{table}

\begin{figure}[thpb]
      \centering
      \includegraphics[width=.9\columnwidth]{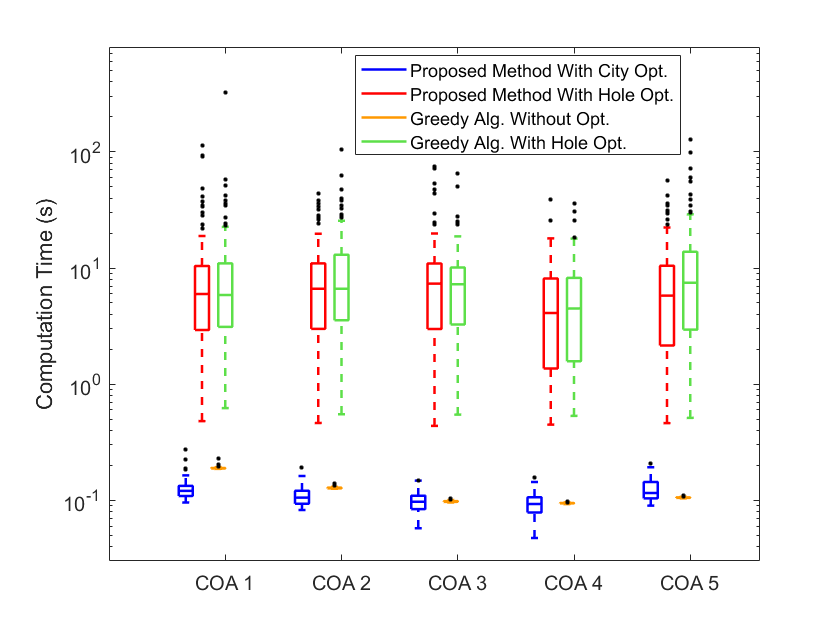}
      \caption{Computation times for all the experiments using the proposed method and greedy method, grouped by COA case.}
      \label{comp_time}
   \end{figure}

{\bf{Computation time:~}} Figure \ref{comp_time} compares the computation times of the four methods. We observe that the computation times for our method with city optimization are lower than those of the greedy scheduler [$F(1,998) = 42.3, p = 1.24 \times 10^{-10}$], and our method solves the scheduling problem in 110.2 ms on an average, and has a maximum computation time of
only 272.3 ms. In practice, this would allow an industrial MRS to start drilling almost immediately. The computation time results in Figure \ref{comp_time} are grouped by COA case, and are summarized in Table \ref{comp_time_table}.

\begin{table}[h]
\caption{Summary Statistics of Computation Times}
\label{comp_time_table}
\begin{center}
\begin{tabular}{|c||c||c|}
\hline
Method & Avg. Comp. Time & Max. Comp. Time\\
\hline
Proposed (City Opt) &  110.2 $\pm$ 26.9 ms & 272.3 ms\\
\hline
Proposed (Hole Opt) & 9.77 $\pm$ 15.9 s & 207.3 s\\
\hline
Greedy (No Opt) &  123.1 $\pm$ 35.5 ms & 230.5 ms\\
\hline
Greedy (Hole Opt) &  10.5 $\pm$ 19.2 s & 323.2 s\\
\hline
\end{tabular}
\end{center}
\end{table}

{\bf{Effect of discretization:~}}
The discretization of the leftover tasks into cities makes the initial scheduling and subsequent leftover optimization problems smaller by an order of magnitude (40 cities versus 400 holes). Therefore, it enables significantly lower computation time as compared to optimizing by holes [$F(1,998) = 182.9, p = 0$], and still yields similar schedule efficiencies [$F(1,998) = 0.13, p = 0.717$]. The value of fast leftover optimization is expected to become more pronounced when the size of the problem is increased, as, for example, when we have more than 10 robots on a full-sized wing, where hole-by-hole optimization would become even more computationally taxing.

{\bf{Failures during leftover scheduling:}}
Our method assumes that no failures occur during the execution of the leftover schedule. While this assumption holds true when the amount of leftovers is small, the risk of incurring a failure in the leftover stage grows as the accumulation of failures increases. The possibility of failures occurring during the leftover schedule execution  
can be precluded by performing preventive maintenance just before the execution of the leftover schedule. The limit for the amount of leftover work is then driven by the failure recurrence distribution.

\section{Conclusions}
In this letter, we adapt partition-based task scheduling to the multi-robot task allocation problem for the assembly of aircraft structures. Our method takes advantage of known problem structure for efficient, collision-free task scheduling, and employs a dual-stage schedule execution strategy to handle robot failures. We present a market-based optimization algorithm to help recover the efficiency lost due to the robot failures. Results show that our method produces high schedule efficiencies and favorable computation times as compared to an optimized greedy method, and works well on a physical four robot assembly cell.

Since our methods are adaptable to different manufacturing operations, we believe that our methods provide a promising foundation for general multi-robot aircraft assembly. Further extensions of this method would investigate the possibility of dynamic city definitions for leftover scheduling, as well as the effects of increasing the occurrence frequency and repair time of failures. In the future, we also plan to deal with task ordering constraints, as, for example, those introduced by tool changes to drill holes of different diameters.

\end{document}